\documentclass[11pt,a4paper]{article}
\usepackage[hyperref]{emnlp2018}
\usepackage{times}
\usepackage{latexsym}
\usepackage{tikz}
\usetikzlibrary{decorations.pathreplacing}

\usepackage{graphicx}
\usepackage{url}
\usepackage{hyperref}
\usepackage{amsmath}
\usepackage{amsfonts}

\usepackage{enumitem}
\usepackage{multirow}
\usepackage{multicol}
\usepackage{afterpage}

\aclfinalcopy 

\title{An Analysis of Attention Mechanisms: The Case of \\Word Sense Disambiguation in Neural Machine Translation}

\author{Gongbo Tang$^1$\quad Rico Sennrich$^{2,3}$\quad Joakim Nivre$^1$ \medskip\\
  $^1$Department of Linguistics and Philology, Uppsala University\\
  $^2$School of Informatics, University of Edinburgh\\
  $^3$Institute of Computational Linguistics, University of Zurich\\
  {\tt firstname.lastname@\{lingfil.uu.se, ed.ac.uk\}}}
  \date{}

\begin{document}
\maketitle

\begin{abstract}
Recent work has shown that the encoder-decoder attention mechanisms in neural machine translation (NMT) are different from the word alignment in statistical machine translation. In this paper, we focus on analyzing encoder-decoder attention mechanisms, in the case of word sense disambiguation (WSD) in NMT models. We hypothesize that attention mechanisms pay more attention to context tokens when translating ambiguous words. We explore the attention distribution patterns when translating ambiguous nouns. Counter-intuitively, we find that attention mechanisms are likely to distribute more attention to the ambiguous noun itself rather than context tokens, in comparison to other nouns. We conclude that attention mechanism is not the main mechanism used by NMT models to incorporate contextual information for WSD. The experimental results suggest that NMT models learn to encode contextual information necessary for WSD in the encoder hidden states. For the attention mechanism in Transformer models, we reveal that the first few layers gradually learn to ``align'' source and target tokens and the last few layers learn to extract features from the related but unaligned context tokens. 
\end{abstract}

\section{Introduction}
\label{sec:intro}

Human languages exhibit many different types of ambiguity. Lexical ambiguity refers to the fact that words can have more than one semantic meaning. Dealing with these lexical ambiguities 
is a challenge for various NLP tasks. 
Word sense disambiguation (WSD) is recognizing the correct meaning of an 
ambiguous word, with the help of contextual information. 

In statistical machine translation (SMT) \cite{koehn2003statistical}, a system 
could explicitly take context tokens into account to improve the translation of 
ambiguous words \cite{vickrey2005word}. 
By contrast, in neural machine translation (NMT) 
\cite{kal2013recurrent,sutskever2014sequence,cho2014learning}, 
especially in attentional NMT \cite{bahdanau15joint,luong2015effective}, 
each hidden state incorporates contextual information. 
Hence, NMT models could potentially perform WSD well. 
However, there are no empirical results to indicate that the hidden states encode the contextual information needed for disambiguation. Moreover, 
how the attention mechanism\footnote{Denotes the encoder-decoder attention mechanism in this paper, unless otherwise specified.} deals with ambiguous words is also not known yet. 

In this paper, we focus on the question of how encoder-decoder attention 
mechanisms deal with ambiguous nouns. 
We explore two different attention mechanisms. One is the vanilla one-layer 
attention mechanism \cite{bahdanau15joint,luong2015effective}, and the other 
one is the Transformer attention mechanism \cite{vaswani2017Attention}. 

\newcite{rios2017improving} find that attentional NMT models perform well in 
translating ambiguous words with frequent senses,\footnote{More than 2,000 instances in the 
training set.} while \newcite{liu2017handling} 
show that there are plenty of incorrect translations of ambiguous words. 
In Section \ref{sec:eval}, we evaluate the translations of ambiguous nouns, 
using the test set from \newcite{rios2017improving}. In this setting, 
we expect to get a more accurate picture of the WSD performance of NMT models. 

In Section \ref{sec:ambi}, we present a fine-grained investigation of attention 
distributions of different attention mechanisms. 
We focus on the process of translating the given ambiguous nouns. 
Previous studies \cite{ghader2017what,koehn2017challenges} have shown that 
attention mechanisms learn to pay attention to some unaligned but useful context 
tokens for predictions. Thus, we hypothesize that attention mechanisms distribute 
more attention to context tokens when translating ambiguous nouns, 
compared to when translating other words. 
To test this hypothesis, we compare the attention weight over ambiguous nouns 
with the attention weight over all words and all nouns. 

In Section \ref{sec:analysis}, we first compare the two different attention 
mechanisms. Then, we explore the relation between accuracy and attention 
distributions when translating ambiguous nouns. In the end, 
we investigate the error distributions over frequency. 

Our main findings are summarized as follows: 
\begin{itemize}[noitemsep]
  \item We find that WSD is challenging in NMT, and data sparsity is one of the 
  main issues. 
  \item We show that attention mechanisms prefer to pay more attention to the 
  ambiguous nouns rather than context tokens when translating ambiguous nouns. 
  \item We conclude that attention mechanism is not the main mechanism used by NMT models to incorporate contextual information for WSD. Experimental results suggest that models learn to encode contextual information necessary for WSD in the encoder hidden states. 
  \item We reveal that the attention mechanism in Transformers first gradually 
  learns to extract features from the ``aligned'' source tokens. Then, it 
  learns to capture features from the related but unaligned source context tokens. 
\end{itemize}

\section{Related Work}

Both \newcite{rios2017improving} and \newcite{liu2017handling} propose some 
techniques to improve the translation of ambiguous words. 
\newcite{rios2017improving} use sense embeddings and lexical chains as additional 
input features. \newcite{liu2017handling} introduce an additional context vector.
There is an apparent difference in evaluation between these two studies. 
\newcite{rios2017improving} design a constrained WSD task. 
They create well-designed test sets to evaluate the performance of NMT models 
in distinguishing different senses of ambiguous words, 
rather than evaluating the translations of ambiguous words directly. 
By contrast, \newcite{liu2017handling} evaluate the translations of ambiguous words 
but on a common test set. 
Scoring the contrastive translations is not evaluating the real output of NMT models. 
In this paper, we directly evaluate the translations  generated by NMT models, 
using \textit{ContraWSD} as the test set. 

In NMT, the encoder may encode contextual information into the hidden states. 
\newcite{marvin2018exploring} explore the ability of hidden states at different 
encoder layers in WSD, while we focus on exploring the attention mechanisms 
that connect the encoder and the decoder. 

\newcite{koehn2017challenges} and \newcite{ghader2017what} investigate 
the relation between attention mechanisms and the traditional word alignment. 
They find that attention mechanisms not only pay attention to the aligned 
source tokens but also distribute attention to some unaligned source tokens. 
In this paper, we perform a more fine-grained investigation 
of attention mechanisms, focusing on the task of translating ambiguous nouns. 
We also explore the advanced attention mechanisms in Transformer models 
\cite{vaswani2017Attention}. 

The encoder-decoder attention mechanisms differ in NMT models. 
\newcite{Tang2018why} evaluate different NMT models, but 
focusing on NMT architectures. 
\newcite{Tang2018evaluation,Domhan2018how} compare different attention mechanisms. 
However, there is no detailed analysis on attention mechanisms. 

In this paper, we mainly investigate the encoder-decoder attention mechanisms. 
More specifically, we explore how attention mechanisms work when translating ambiguous nouns.

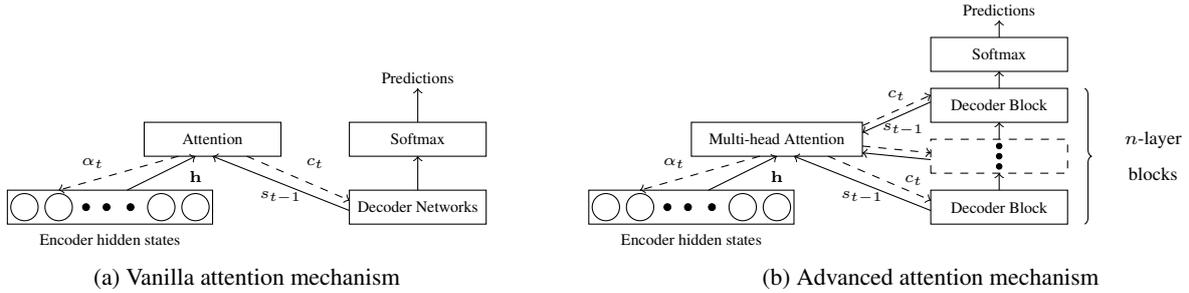
\begin{figure*}[htbp]
\centering
\begin{tikzpicture}[scale=0.45]
\draw (-2,0) rectangle (4,1); 
\draw (-1.5,0.5) circle [radius=0.4] (-0.5,0.5) circle [radius=0.4] (2.5,0.5) circle [radius=0.4] (3.5,0.5) circle [radius=0.4]; 
\draw [fill] (0.3,0.5) circle [radius=0.1] (0.9,0.5) circle [radius=0.1] (1.6,0.5) circle [radius=0.1];
\draw [->] [dashed] (3,2) -- (-0.5,1); \draw [->] (1.5,1) -- (3.5,2);
\node at (0.5,1.8) {\tiny $\alpha_{t}$};\node at (3.5,1.4) {\tiny $\bf h$};
\node [below] at (1,0.0) {\tiny Encoder hidden states};
\draw (2,2) rectangle (6,3); \node at (4,2.5) {\tiny Attention};
\draw (8,0) rectangle (12,1) (8,2) rectangle (12,3) ; 
\node at (10,0.5) {\tiny Decoder Networks};
\draw [->] (10,1) -- (10,2);
\draw [->] (10,3) -- (10,4);  
\node at (10,2.5) {\tiny Softmax}; \node at (10,4.3) {\tiny Predictions};
\draw [->] [dashed] (5,2) -- (8,0.7);  \draw [->] (8,0.4) -- (4,2);
\node at (6,0.8) {\tiny $s_{t-1}$};\node at (7,1.8) {\tiny $c_{t}$};

\node [below] at (5,-1) {\small (a) Vanilla attention mechanism};

\draw (15,0) rectangle (21,1); 
\draw (15.5,0.5) circle [radius=0.4] (16.5,0.5) circle [radius=0.4] (19.5,0.5) circle [radius=0.4] (20.5,0.5) circle [radius=0.4]; 
\draw [fill] (17.3,0.5) circle [radius=0.1] (17.9,0.5) circle [radius=0.1] (18.6,0.5) circle [radius=0.1];
\draw [->] [dashed] (20,2) -- (16.5,1);  \draw [->] (18.5,1) -- (20.5,2);
\node at (17.5,1.8) {\tiny $\alpha_{t}$};\node at (20.5,1.4) {\tiny $\bf h$};
\node [below] at (18,0.0) {\tiny Encoder hidden states};
\draw (18,2) rectangle (23,3); \node at (20.5,2.5) {\tiny Multi-head Attention};
\draw (25,0) rectangle (29,1) (25,3) rectangle (29,4) (25,4.5) rectangle (29,5.5); 
\draw [dashed] (25,1.5) rectangle (29,2.5);
\draw [fill] (27,1.7) circle [radius=0.08] (27,2) circle [radius=0.08](27,2.3) circle [radius=0.08];
\node at (27,0.5) {\tiny Decoder Block};\node at (27,3.5) {\tiny Decoder Block};
\draw [->] (27,1) -- (27,1.5); \draw [->] (27,2.5) -- (27,3);
\draw [->] (27,4) -- (27,4.5); \draw [->] (27,5.5) -- (27,6);  
\node at (27,5) {\tiny Softmax}; \node at (27,6.3) {\tiny Predictions};
\draw [->] [dashed] (22,2) -- (25,0.7);  \draw [->] (25,0.4) -- (21,2);
\draw [->] [dashed] (23,2.3) -- (25,2.1);  \draw [->] (25,1.9) -- (23,2.1);
\draw [->] [dashed] (23,2.9) -- (25,3.8);  \draw [->] (25,3.6) -- (23,2.7);
\node at (23,0.8) {\tiny $s_{t-1}$};\node at (24.5,1.3) {\tiny $c_{t}$};
\node at (24.2,2.8) {\tiny $s_{t-1}$};\node at (24,3.8) {\tiny $c_{t}$};
\draw [decorate, decoration={brace,mirror}] (29.5,0)--(29.5,4);
\node [align=center] at (31.5,2) {\scriptsize $n$-layer \\ \scriptsize blocks};

\node [below] at (25,-1) {\small (b) Advanced attention mechanism};
\end{tikzpicture}        
\caption{Different attention mechanisms between encoders and decoders in NMT.}
\label{fig:attention_mechanisms}
\end{figure*}

\section{Background}
\label{sec:background}

\subsection{Attention Mechanisms} 
\label{sub:attention_mechanisms}

Attention mechanisms were initially proposed to learn the alignment between source 
and target tokens by \newcite{bahdanau15joint} and \newcite{luong2015effective}, 
in order to improve the performance of NMT. 
However, attention mechanisms are different from the traditional word alignment in SMT 
which learns the hard alignment between source and target tokens.  
Attention mechanisms learn to extract features from all the source tokens 
when generating a target token. 
They assign weights to all the hidden states of source tokens. 
The more related hidden states are assigned larger weights. 
Then attention mechanisms feed a \textit{context vector} $c_{t}$, which is 
extracted from the encoder, into the decoder for target-side predictions. 

We use $\bf h$ to represent the hidden state set $\{h_{1}, h_{2}, \cdots, h_{n}\}$ 
in the encoder, where $n$ is the number of source-side tokens. 
Then $c_{t}$ is computed by Equation \ref{eq:context-vector}: 
\begin{equation} \label{eq:context-vector}
c_{t} = \alpha_{t}\bf h 
\end{equation} 
where $\alpha_{t}$ is the attention vector at time step $t$. 
$\alpha_{t}$ is a normalized distribution of a score  computed by the 
hidden state set $\bf h$ and the decoder state $s_{t-1}$, 
as described by Equation \ref{eq:att-vector}: 
\begin{equation} \label{eq:att-vector}
a_{t} = \mathit{softmax}(\mathit{score}(s_{t-1}, \bf h))
\end{equation} 
There are different $\mathit{score}()$ functions to compute the \textit{attention vector} $a_{t}$, 
including multi-layer perceptron (MLP), dot product, multi-head attention, etc. 
In this paper, the vanilla attention mechanism employs MLP. The advanced 
attention mechanism applies multi-head attention with scaled dot product, which is the 
same as the attention mechanism in Transformer \cite{vaswani2017Attention}.

Figure \ref{fig:attention_mechanisms} illustrates different attention mechanisms. 
In vanilla attention mechanisms \cite{bahdanau15joint,luong2015effective}, 
the \textit{context vector} $c_{t}$ is only fed into the first layer 
of the decoder networks. Then the single- or multi-layer decoder networks compute 
from bottom to top to predict target tokens. 
The vanilla attention mechanisms can only extract the source-side features once, 
which may be insufficient. Therefore, \newcite{gehring2017convolutional} and 
\newcite{vaswani2017Attention} feed a context vector into each decoder layer. 
The higher layer could take the result of the previous layer 
into account when computing the new attention. 
More recently, \newcite{Domhan2018how} has shown that multi-layer attention 
is crucial in NMT models. 
Moreover, \newcite{vaswani2017Attention} also propose the multi-head attention 
mechanism. In contrast to the single-head attention, 
there are multiple attention functions which compute the attention from the 
linearly projected vectors in parallel. Then, the context vectors 
from all the heads are concatenated and fed into the decoder networks.

\subsection{\textit{ContraWSD}}
\label{sub:contrawsd}

\textit{ContraWSD}\footnote{\href{https://github.com/a-rios/ContraWSD}{
https://github.com/a-rios/ContraWSD}} from \newcite{rios2017improving} 
consists of contrastive translation sets where the human 
reference translations are paired with one or more contrastive variants. 
Given an ambiguous word in the source sentence, the correct translation is replaced 
by an incorrect translation corresponding to another meaning of the ambiguous word. 
For example, in a case where the English word `line' is the correct translation 
of the German source word `Schlange', \textit{ContraWSD} replaces `line' 
with other translations of `Schlange', such as 
`snake' or  `serpent', to generate contrastive translations. 
To evaluate the performance on disambiguation, contrastive translations are designed 
not to be easily identified as incorrect based on grammatical and phonological features.

\textit{ContraWSD} is extracted from a large amount of balanced parallel text. 
It contains 84 different German word senses. 
It has 7,200 German$\rightarrow$English lexical ambiguities and each lexical 
ambiguity instance has 3.5 contrastive translations on average. 
All the ambiguous words are nouns so that the WSD is 
not simply based on syntactic context.

\section{Evaluation}
\label{sec:eval}

Instead of using NMT models to score the contrastive translations, 
we use NMT models to translate source sentences and evaluate the translations 
of the ambiguous nouns directly. 
We evaluate two popular NMT models with different attention mechanisms. 
One is \textit{RNNS2S} with the vanilla attention mechanism, and the other is 
\textit{Transformer} with the advanced attention mechanism. 

We apply \textit{fast-align} \cite{dyer2013fast} to get the aligned translations 
of ambiguous nouns. To achieve better alignment, we run \textit{fast-align} 
on both training data and test data which includes reference translations 
and generated translations. However, for some ambiguous nouns, 
there is no alignment. We call these ambiguous nouns  \emph{filtered}. 

There are multiple reference translations for each ambiguous noun in 
\textit{ContraWSD}. We additionally add their synonyms\footnote{Synonyms from 
WordNet \cite{miller1995wordnet}} into the reference translations as well.  
The non-reference translations are crawled from the 
Internet\footnote{\href{https://www.linguee.com/german-english}
{https://www.linguee.com/german-english}}.

In addition to the \emph{filtered} nouns, the translations of the ambiguous nouns are 
classified into six groups, depending on which class (references, incorrect senses, 
no translation) the translations at aligned/unaligned positions belong to, 
as described in Table \ref{table-trans-group}. For instance, 
in \textit{C3}, there is neither a correct nor an incorrect sense at the aligned position. 
However, there is a reference translation at an unaligned position. 

\begin{table}[htbp]
\begin{centering}
\resizebox{\columnwidth}{!}{
\begin{tabular}{|c|c|c|c|c|c|c|}
\hline \multirow{2}{*}{Group} &\multicolumn{3}{c|}{Aligned} & \multicolumn{3}{c|}{Unaligned}\\
\cline{2-7} \ &Ref.&Incor.&No&Ref.&Incor.&No\\
\hline \textit{C1} &$\surd$&&&&&\\
\hline \textit{C2} &&$\surd$&&$\surd$&&\\
\hline \textit{W1} &&$\surd$&&&$\surd$&$\surd$\\
\hline \textit{C3} &&&$\surd$&$\surd$&&\\
\hline \textit{W2} &&&$\surd$&&$\surd$&\\
\hline \textit{Drop} &&&$\surd$&&&$\surd$\\
\hline
\end{tabular}}
\caption{\label{table-trans-group} Different groups of translations. 
\textit{Ref.} denotes the reference translations. 
\textit{Incor.} represents the incorrect senses. 
\textit{No} means that there is neither a correct nor an incorrect sense of the ambiguous noun. 
$\surd$ indicates that the translations belong to 
the reference translations or incorrect senses or neither. }
\end{centering}
\end{table}

\noindent
Since the alignment learnt by \textit{fast-align} is not perfect, we also consider 
the translations at unaligned positions. 
All the translations in \textit{C1, C2, C3} groups are viewed as correct translations. 
Thus, the accuracy of an NMT model on this test set is the amount of translations 
in Group \textit{C1, C2, C3}, divided by the sum of ambiguous noun instances. 
Formally, $\mathit{Accuracy}=(C1+C2+C3)/(C1+C2+W1+C3+W2+\mathit{Drop}+\mathit{Filtered})$, 
where $C1$, $C2$, $W1$, $C3$, $W2$, $\mathit{Drop}$, and $\mathit{Filtered}$ are the amount of 
translations in each group.

\subsection{Experimental Settings}
We use the \textit{Sockeye} \cite{Hieber2017sockeye} toolkit, 
which is based on MXNet \cite{chen2015mxnet}, to train models. 
In addition, we have extended \textit{Sockeye} to output the distributions of 
encoder-decoder attention in Transformer models, from 
different attention heads and different attention layers. 

All the models are trained with 2 GPUs. 
During training, each mini-batch contains 4096 tokens. 
A model checkpoint is saved every 4,000 updates. 
We use \textit{Adam} \cite{Kingma2014AdamAM} as the optimizer. 
The initial learning rate is set to 0.0002. 
If the performance on the validation set has not improved for 8 
checkpoints, the learning rate is multiplied by 0.7. 
We set the early stopping patience to 32 checkpoints. 
All the neural networks have 8 layers. For \textit{RNNS2S}, the 
encoder has 1 bi-directional LSTM and 6 stacked uni-directional LSTMs, 
and the decoder is a stack of 8 uni-directional LSTMs. 
The size of embeddings and hidden states is 512.
We apply layer-normalization and label smoothing (0.1) in all models.
We tie the source and target embeddings. 
The dropout rate of embeddings and Transformer blocks is set to 0.1. 
The dropout rate of RNNs is 0.2. 
The attention mechanism in \textit{Transformer} has 8 heads. 

We use the training data from the WMT17 shared 
task.\footnote{\url{http://www.statmt.org/wmt17/translation-task.html}}
We choose \textit{newstest2013} as the validation set, and use 
\textit{newstest2014} and \textit{newstest2017} as the test sets. 
All the BLEU scores are measured by \textit{SacreBLEU}. 
There are about 5.9 million sentence pairs in the training set after 
preprocessing with Moses scripts. 
We learn a joint BPE model with 32,000 subword units \cite{sennrich16sub}. 
There are 6,330 sentences left after filtering the sentences with 
segmented ambiguous nouns. 
We employ the models that have the best perplexity on the validation set 
for the evaluation.

\begin{table*}[htbp]
\begin{center}
\begin{tabular}{|c||c|c||c|c|c|c|c|c|c||c||c|}
\hline Model&\textit{2014}&\textit{2017}&\textit{C1}&\textit{C2}&\textit{W1}&\textit{C3}&\textit{W2}&\textit{Drop}&\textit{Filtered}&\textit{Acc.}&\textit{Score}\\
\hline \textit{RNNS2S} & 23.3& 25.1&4,560 & 187 & 863 & 81 & 31 & 333&275&76.27&84.01\\
\hline \textit{Transformer} &26.7&27.5&4,982 & 140 & 599 & 85 & 23 & 308&193&82.26&90.34\\
\hline
\end{tabular}
\caption{\label{table-eval-result} Evaluation results of NMT models and the 
distributions of translations. \textit{2014} and \textit{2017} denote the 
BLEU scores on \textit{newstest2014} and \textit{newstest2017}, 
\textit{Acc.} (in \%) is short for accuracy. 
\textit{Score} (in \%) is the accuracy using NMT models to score contrastive translation pairs. 
\textit{Filtered} is the amount of translations that there is no learnt alignment 
for the ambiguous nouns. }
\end{center}
\end{table*}

\subsection{Results}

Table \ref{table-eval-result} gives the performance of NMT models on 
\textit{newstest}s and \textit{ContraWSD}. 
The detailed translation distributions over different groups are also provided. 
\textit{Transformer} is much better than \textit{RNNS2S} in both \textit{newstest}s 
and \textit{ContraWSD}. 
Compared to the accuracy of scoring contrastive translation pairs (\textit{Score}), 
the accuracy of evaluating the translations (\textit{Acc.}) is apparently lower. 

There are 8--10\% of ambiguous nouns belonging to \textit{Drop} and 
\textit{Filtered} for both models. 
We manually checked the translations of sentences with these ambiguous nouns and 
found that 250 and 206 ambiguous nouns (41\%) are translated correctly by 
\textit{RNNS2S} and \textit{Transformer}, respectively. Our automatic classification failed for two reasons.
On the one hand, because the models are trained at subword-level, there are a lot of subwords in the translations. 
The correctly generated translations are subword sequences, and not all 
the subwords (sometimes even no subword) are aligned to the ambiguous nouns by \textit{fast-align}. 
On the other hand, the reference translations are all nouns. If the translations 
are verbs or variants, they are not recognized. 
If we move these translations into \textit{C1}, the accuracy of the two NMT models 
will be improved from 76.27\% to 80.22\%, and from 82.26\% to 85.51\%, respectively. 
Thus, attentional NMT models are good at sense disambiguation 
in German$\rightarrow$English, but there is much room for improvement as well.

\section{Ambiguous Nouns in Attentional NMT}
\label{sec:ambi}

\newcite{ghader2017what} show that there are different attention patterns 
for words of different part-of-speech (POS) tags, which sheds light on 
interpreting attention mechanisms. In this section, 
we investigate the attention distributions over source-side ambiguous nouns. 

\subsection{Hypothesis and Tests}
  \label{sub:hypothesis_and_tests}

Attention mechanisms not only pay attention to the hidden states at aligned 
positions but also distribute attention to the hidden states at unaligned 
positions. The hidden states at unaligned positions can influence the generation 
of the current token. In general, NLP models disambiguate ambiguous words 
by means of context words. Thus, for ambiguous nouns, we hypothesize that 
attention mechanisms distribute more attention to context tokens for disambiguation. 

We test our hypothesis via two different comparisons. 
We use $w_{ambi}$ to denote the average attention weight over the ambiguous nouns 
and employ $w_{nouns}$ to represent the average attention weight over all nouns\footnote{We use the \href{http://www.cis.uni-muenchen.de/~schmid/tools/TreeTagger/}
{TreeTagger} \cite{schmid1999treetagger} to tag German.} 
(including the ambiguous nouns), while $w_{tokens}$ denotes the average attention weight 
over all tokens.\footnote{Subword tokens are excluded, which account for 32\%.}
We first compare $w_{ambi}$ with $w_{tokens}$. 
As nouns have a more concentrated attention distribution than other word types 
\cite{ghader2017what}, we then compare $w_{ambi}$ with $w_{nouns}$. 
If $w_{ambi}$ is the smallest, it supports our hypothesis. 

The NMT models we evaluated are trained at subword-level. 
When we compute the attention distributions, we only consider the ambiguous nouns 
that are not segmented into subwords. To some extent, we therefore conduct an analysis of frequent tokens.
We employ the alignment learnt by \textit{fast-align} to find the step of translating 
the current source token. 

Given the attention distribution matrix $M\in\mathbb{R}^{l_{s}*l_{t}}$ of 
a sentence translation, $l_{t}$ represents the length of the target sentence, while
$l_{s}$ denotes the length of the source sentence. 
Each column is the attention distribution over all the source tokens when 
generating the current target token. 
Each row is the attention distribution over the current source token 
at all the translation steps. 
$w$ represents the attention weight over any tokens. 
If the $i$th source token is aligned to the $j$th target token, 
then $w=\left[M\right]_{ij}$. If a token is aligned to more than one token, 
we choose the largest attention weight as $w$.\footnote{A source token may be 
aligned to a set/subset of subword sequences, but the attention mechanism only 
assigns the corresponding weight to one of the subwords. 
We select the maximal weight rather than the average weight.}

As for Transformer attention mechanisms, there are multiple layers, and 
each layer has multiple heads. We maximize the attention weights in 
different heads to represent the attention distribution matrix for 
each attention layer.\footnote{We visualize both the maximal and average 
attention weights. We find that maximal attention weights are 
more representative in feature extraction.} 
We first compute $w_{ambi}$, $w_{nouns}$, and $w_{tokens}$ for each attention layer. 
Then we average these weights. 

\subsection{Results}
\label{sub:results}

As Table \ref{table-ambi-weights} shows, $w_{ambi}$ is substantially larger than 
$w_{tokens}$ in both two models. Even though $w_{nouns}$ is much larger compared 
to $w_{tokens}$, $w_{ambi}$ is still greater than $w_{nouns}$, especially in 
\textit{Transformer}. This result is against our hypothesis. That is to say, 
attention mechanisms do not distribute more attention to context tokens 
when translating an ambiguous noun. 
Instead, attention mechanisms pay more attention to the ambiguous noun itself. 
We assume that the contextual information has already been encoded into the 
hidden states by the encoder, and attention mechanisms do not learn which 
source words are useful for WSD. 

\begin{table}[htbp]
\begin{center}
\begin{tabular}{|c|c|c|c|}
\hline Model & $w_{ambi}$&$w_{tokens}$&$w_{nouns}$\\ 
\hline \textit{RNNS2S} & 0.63&0.48&0.62\\
\hline \textit{Transformer} &0.74&0.57&0.69\\
\hline
\end{tabular}
\caption{\label{table-ambi-weights} Average attention weights over ambiguous nouns, 
non-subword tokens, and nouns.}
\end{center}
\end{table}

\begin{figure}[htbp]
\centering
        \includegraphics[totalheight=4cm]{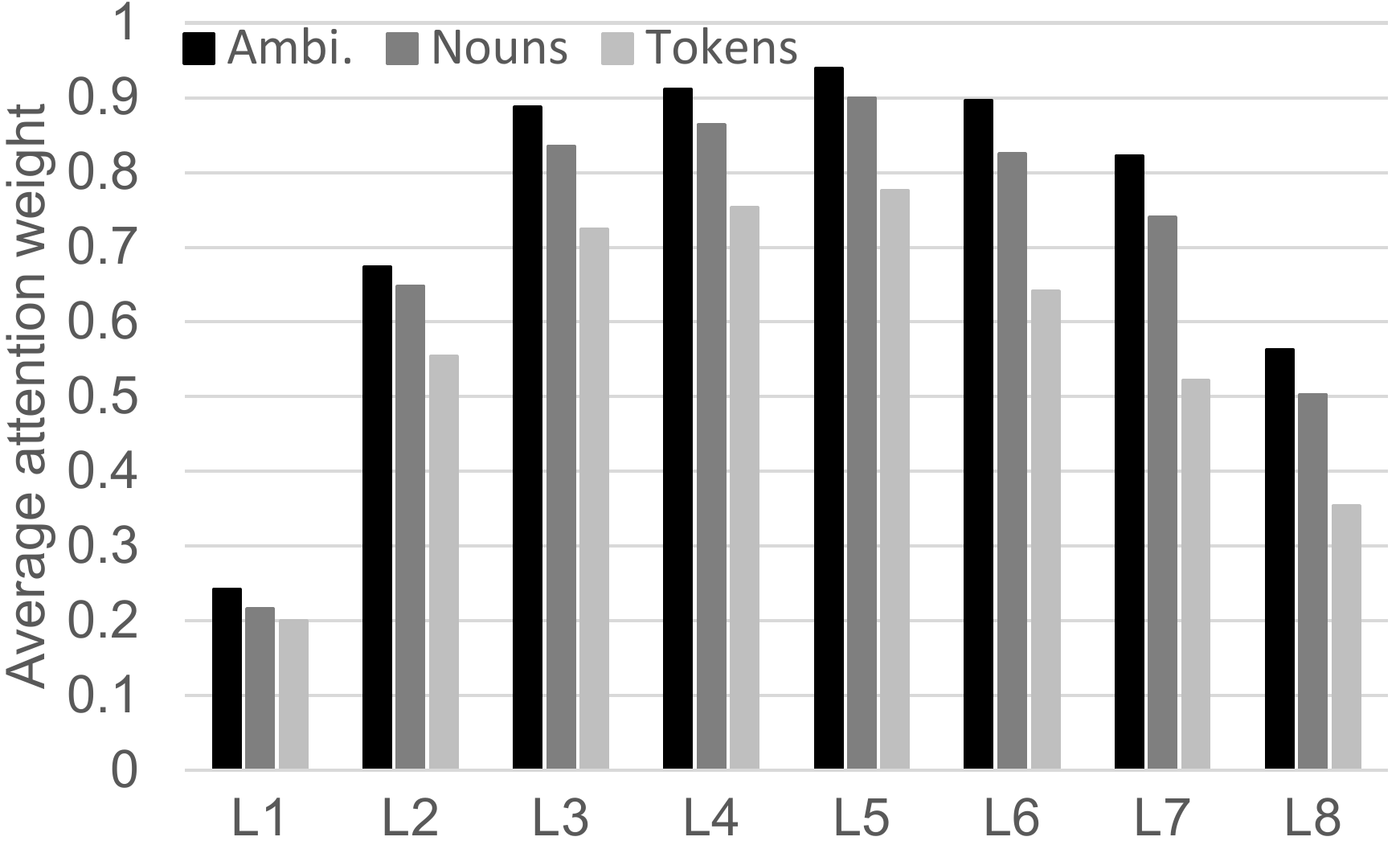}
    \caption{Average attention weights of  ambiguous nouns, nouns, and non-subword 
    tokens in different Transformer attention layers.}
    \label{fig:weights-transformer}
\end{figure}

\noindent
Figure \ref{fig:weights-transformer} demonstrates the average attention weights of 
the ambiguous nouns, nouns, and non-subword tokens in different Transformer attention 
layers. In each attention layer, $w_{ambi}$ is always the largest attention weight. 
It is very interesting that the attention weights keep increasing at lower layers 
and achieve the largest weight at Layer 5. Then $w_{tokens}$ decreases steadily, 
while $w_{ambi}$ and $w_{nouns}$ have a distinct drop in the final attention layer. 
We also re-train a model with 6 attention layers, and we get a figure with 
the same pattern, but the largest attention weights appear at Layer 4. 
We will give a further analysis of Transformer attention mechanisms in Section 
\ref{sub:vanilla_advanced}. 

\section{Analysis}
\label{sec:analysis}

We first give our analysis of the two different attention mechanisms based on the 
attention distributions and visualizations. 
Then, we explore the relation between translation accuracy and attention weight over 
the ambiguous nouns. 
In the end, we provide the error distributions over frequency. 

\begin{figure*}[htbp]
\centering
        \includegraphics[totalheight=5.8cm]{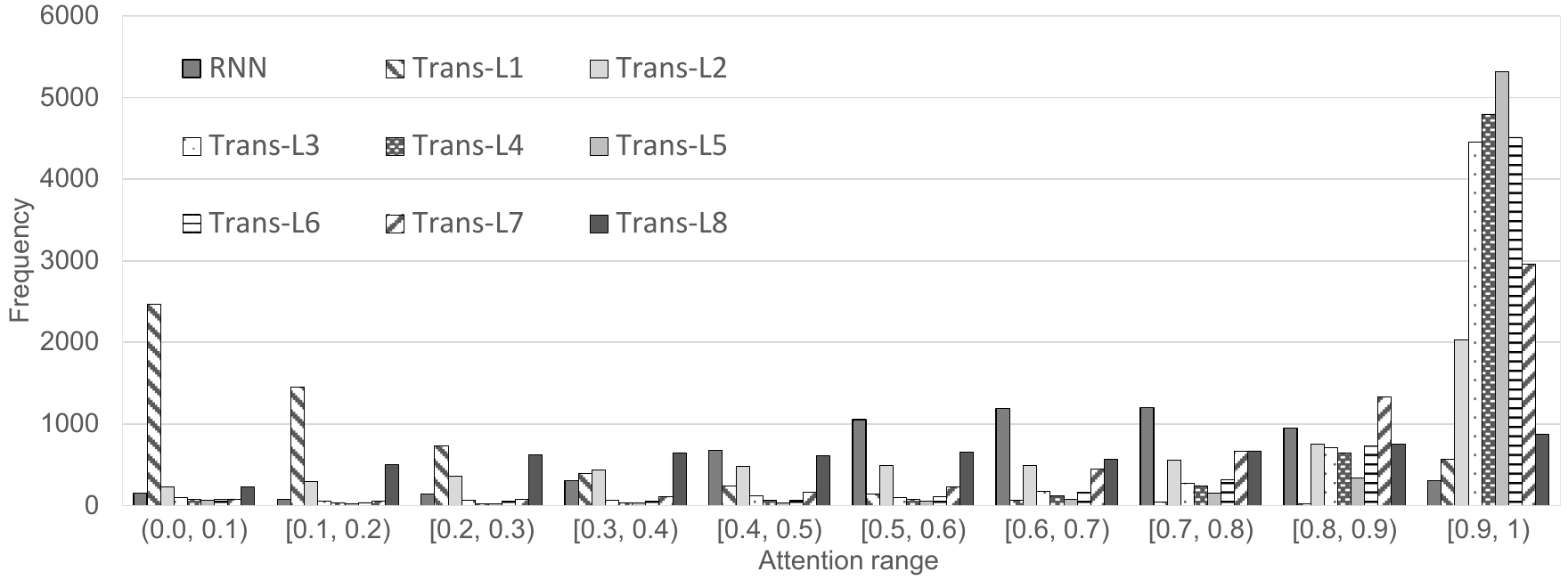}
    \caption{Attention distributions for translating ambiguous nouns from different models.
     \textit{Trans-L3} denotes the third attention layer in the Transformer model. }
    \label{fig:weights-distribution}
\end{figure*}

\subsection{Vanilla Attention vs. Advanced Attention} 
\label{sub:vanilla_advanced}

As Table \ref{table-eval-result} shows, the Transformer model with advanced 
attention mechanisms is distinctly better than the RNN model with vanilla attention 
mechanisms. Even though there are differences in the encoder and decoder networks, 
we focus on the comparison between these two attention mechanisms. Moreover, 
there is no existing empirical interpretation of the advanced attention mechanisms. 

Figure \ref{fig:weights-distribution} demonstrates the attention distributions of 
different models when translating ambiguous nouns. 
For the vanilla attention mechanism in the RNN model, most of the attention weights 
are relatively uniformly distributed in $[0.5, 0.9)$. While the patterns in 
advanced attention mechanisms are completely different. 
In the first layer, most of the attention weights are smaller than $0.1$. 
The larger attention weights, the fewer instances, except when the weight 
is larger than $0.9$. In the following layers, the attention weights are getting 
more and more concentrated in $[0.9, 1)$ until the fifth layer. 
After the fifth layer, the amount in $[0.9, 1)$ decreases dramatically. 
We hypothesize that the first few layers are learning the ``alignment'' gradually. 
When attention mechanisms finish the ``alignment'' learning, they start to capture 
contextual features from the related but unaligned context tokens. 
In the last layer, the attention is almost equally distributed over all the 
attention ranges except $(0, 0.1)$. That is to say, for some ambiguous nouns, 
the weights are large. For the other ambiguous nouns, the weights are small. 
It indicates that there is no clear attention distribution pattern over 
ambiguous nouns in the last layer. 

\begin{figure}[htbp]
\centering
        \includegraphics[totalheight=4.0cm]{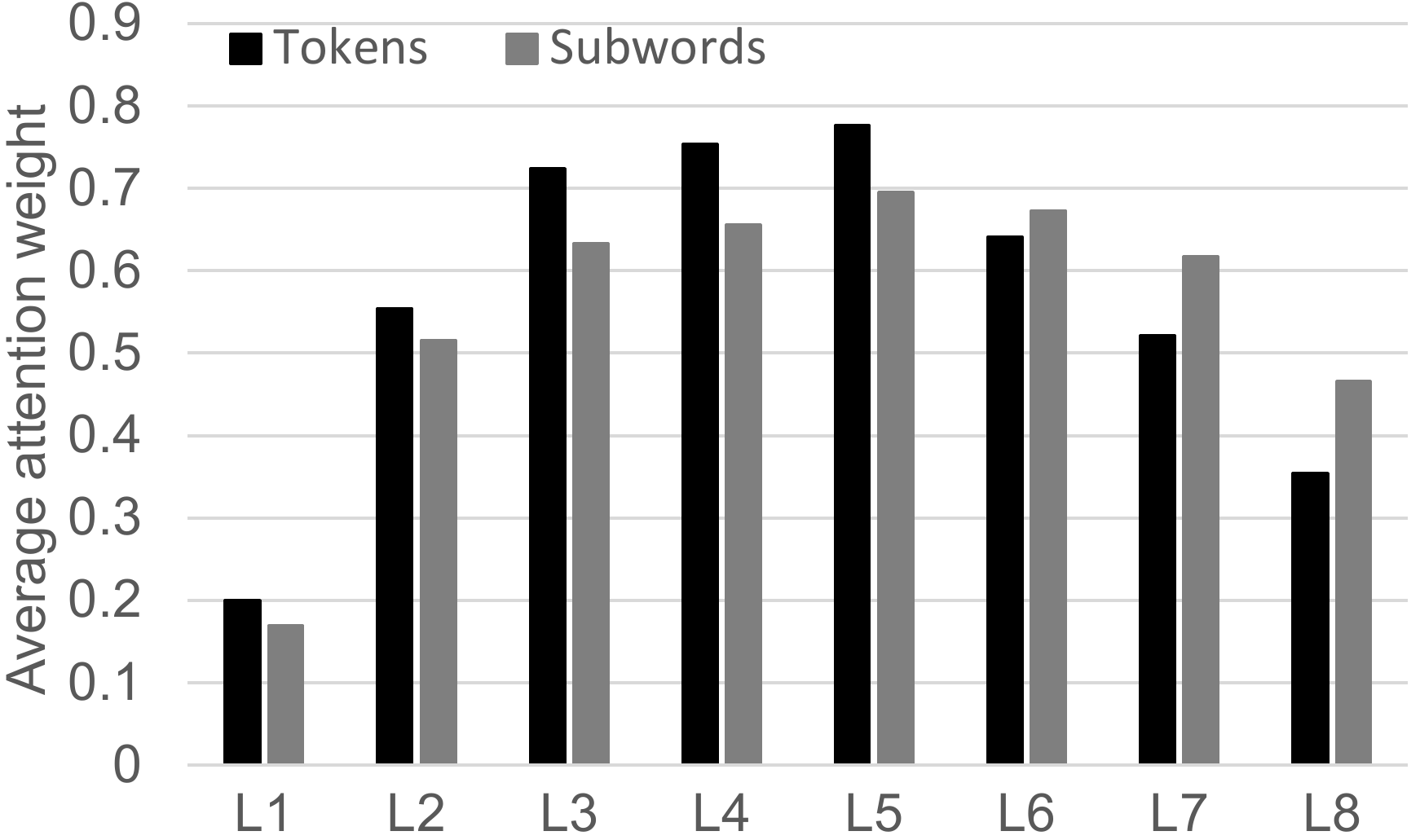}
    \caption{Average attention weights of non-subword tokens and subwords 
    in different Transformer attention layers.}
    \label{fig:token-subword}
\end{figure}

Figure \ref{fig:token-subword} shows the average attention weights over word 
tokens and subword tokens ($w_{subwords}$). 
In the first five layers, $w_{subwords}$ is clearly lower than $w_{tokens}$ which 
can be taken to show that attention mechanisms focus on the ``alignment'' of single 
word tokens, while $w_{subwords}$ surpasses $w_{tokens}$ from the sixth layer. 
We conclude that attention mechanisms focus on subwords instead of word tokens. 
Many words are segmented into multiple consecutive subwords and not all the 
subwords are aligned to the expected target tokens. Thus, the pattern over 
subword tokens demonstrates that attention mechanisms are learning to capture 
context-level features.

\afterpage{
\begin{table*}[ht]
\begin{center}
\begin{tabular}{c@{\hskip -0.02in}c@{\hskip -0.02in}c}
\includegraphics[totalheight=3.9cm]{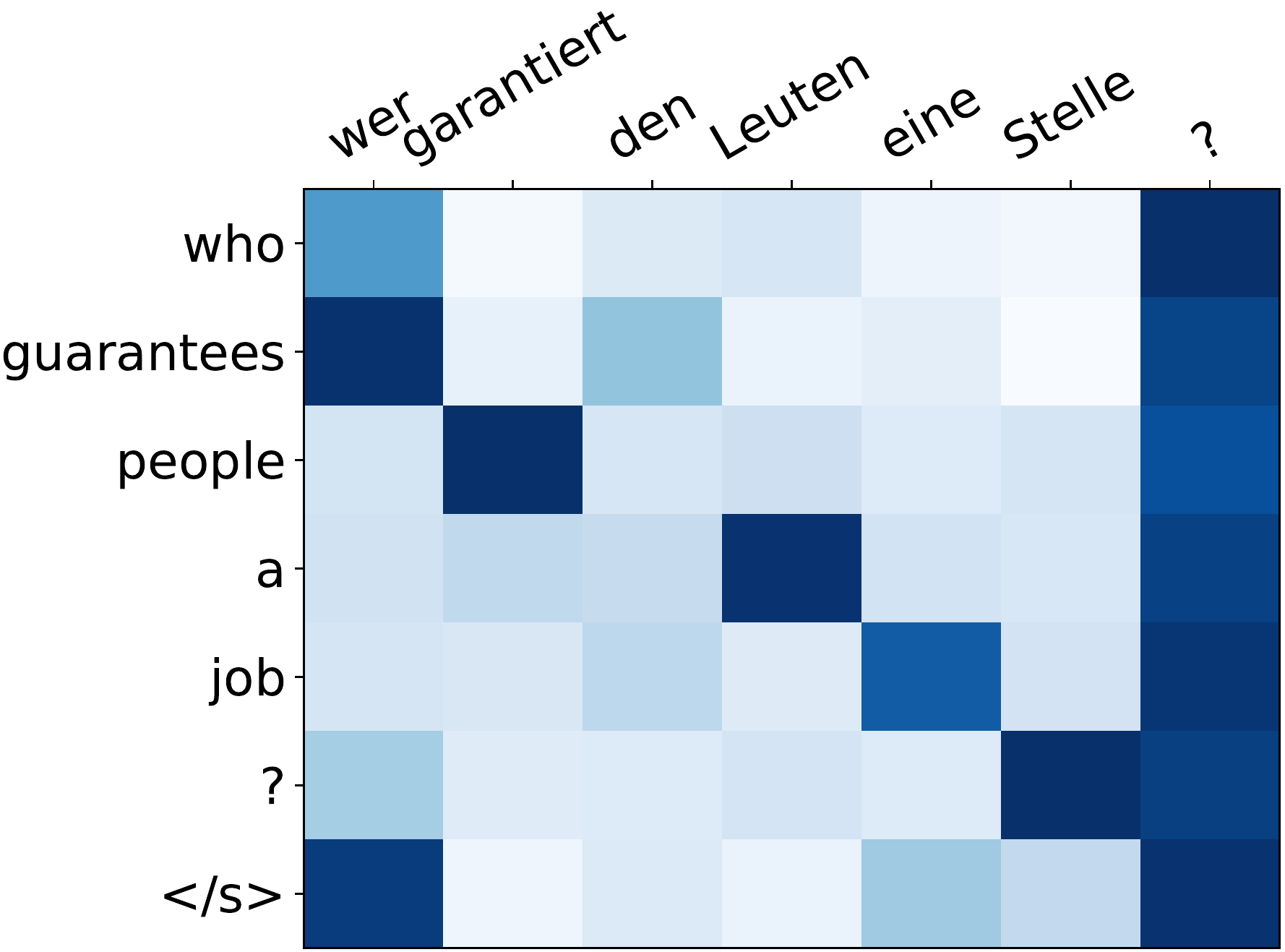} & 
\includegraphics[totalheight=3.9cm]{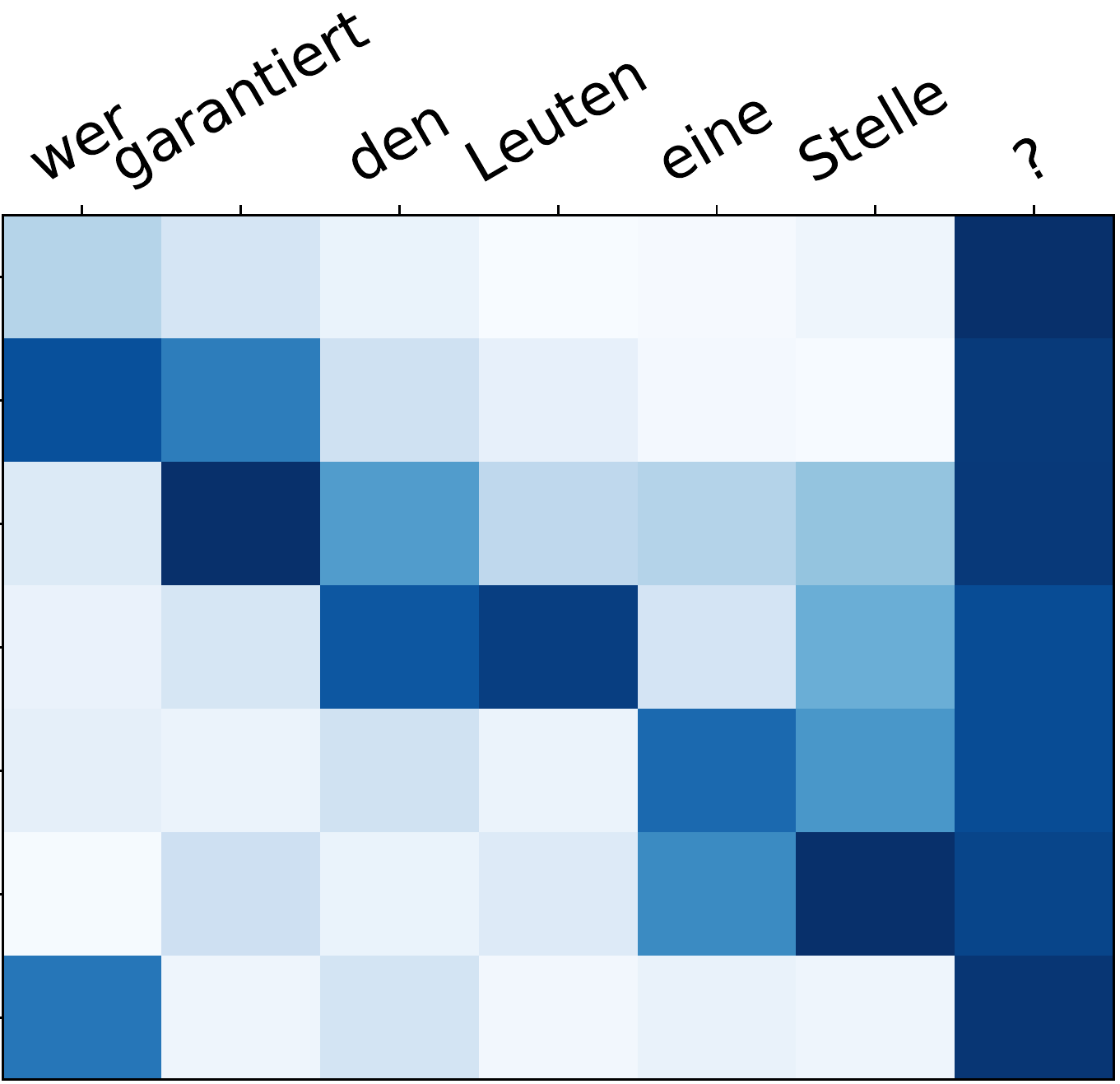} & 
\includegraphics[totalheight=3.9cm]{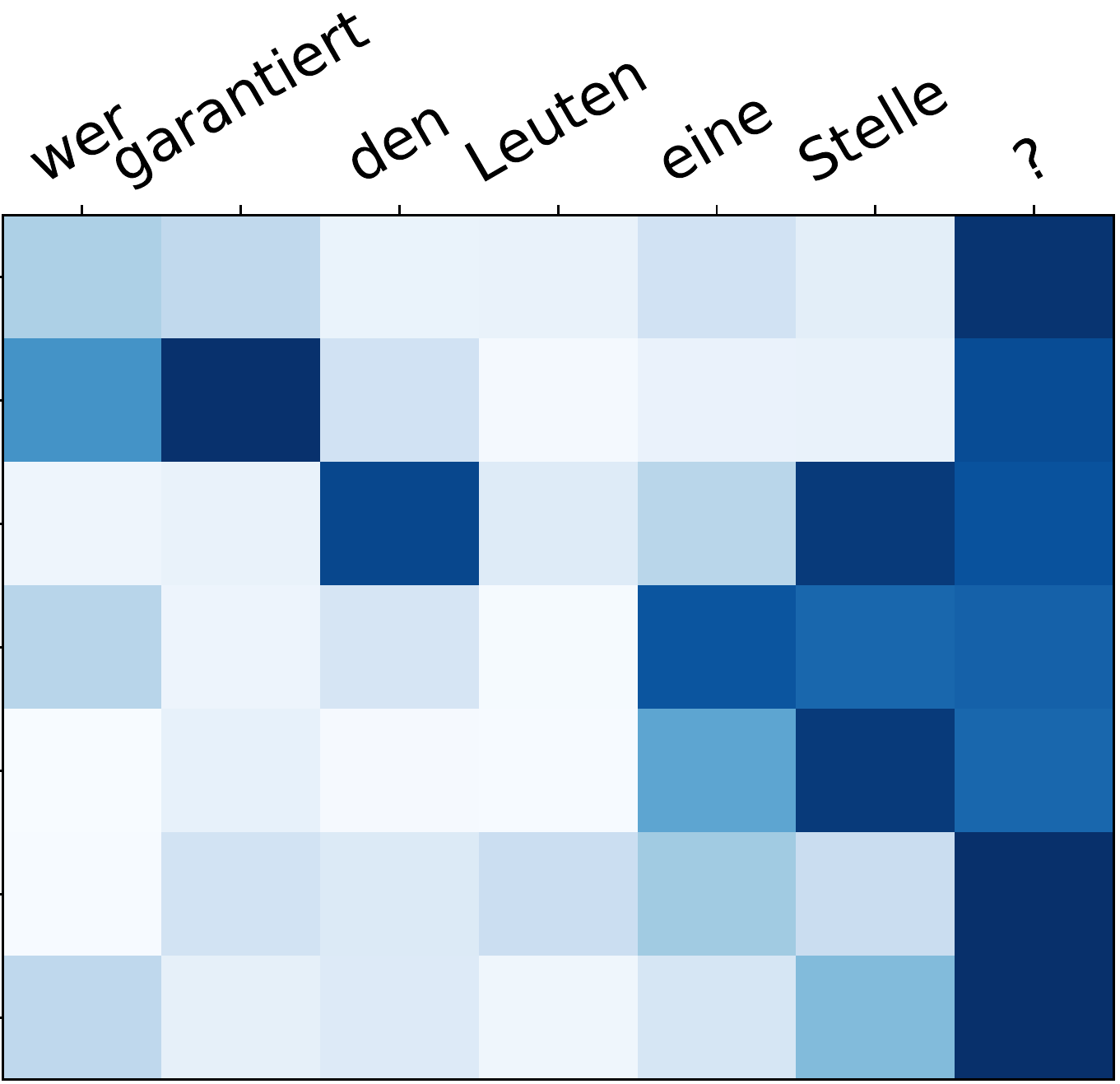} \\
\small (a) Layer 1 & \small (b) Layer 2 & \small(c) Layer 3\\
\includegraphics[totalheight=3.17cm]{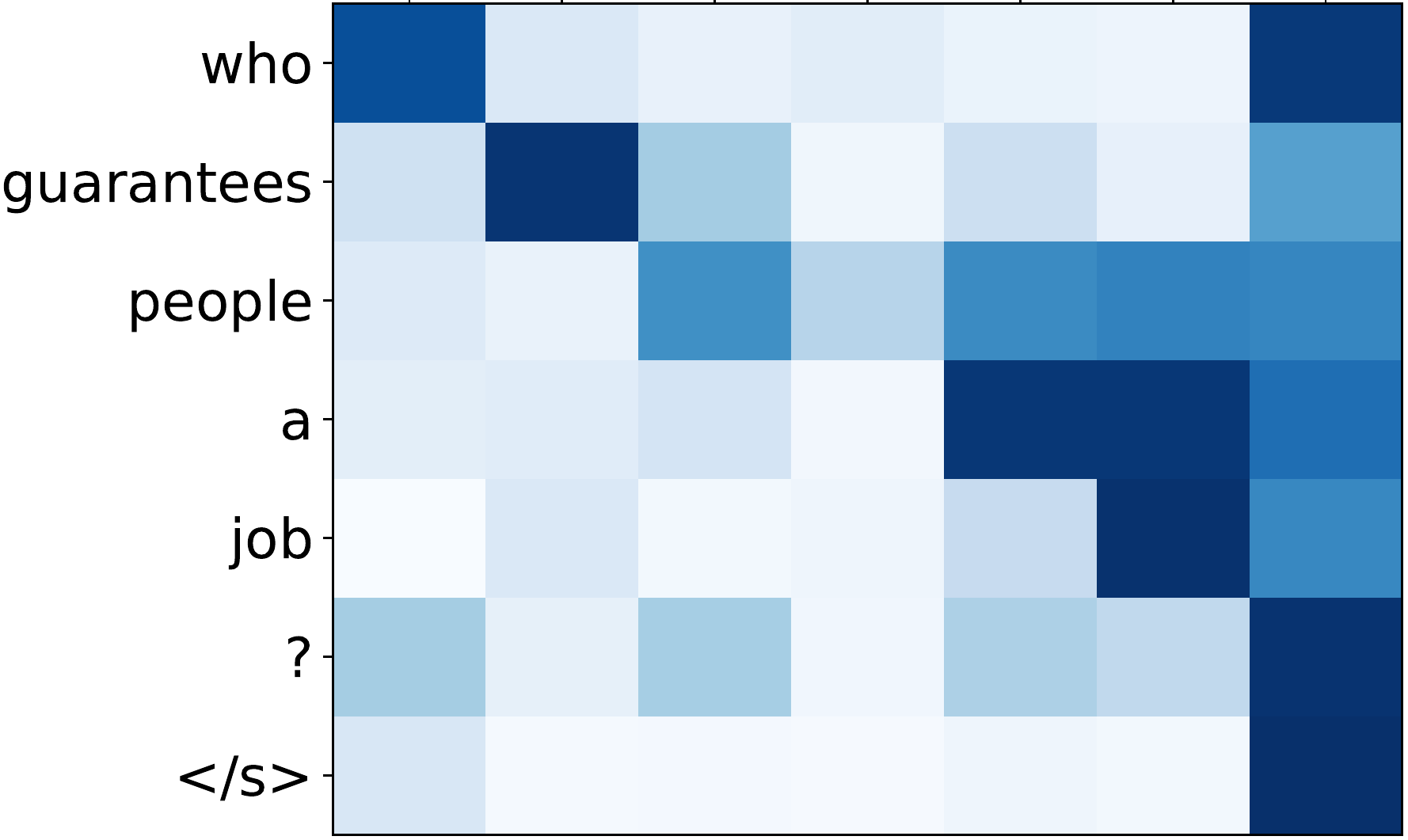} & 
\includegraphics[totalheight=3.17cm]{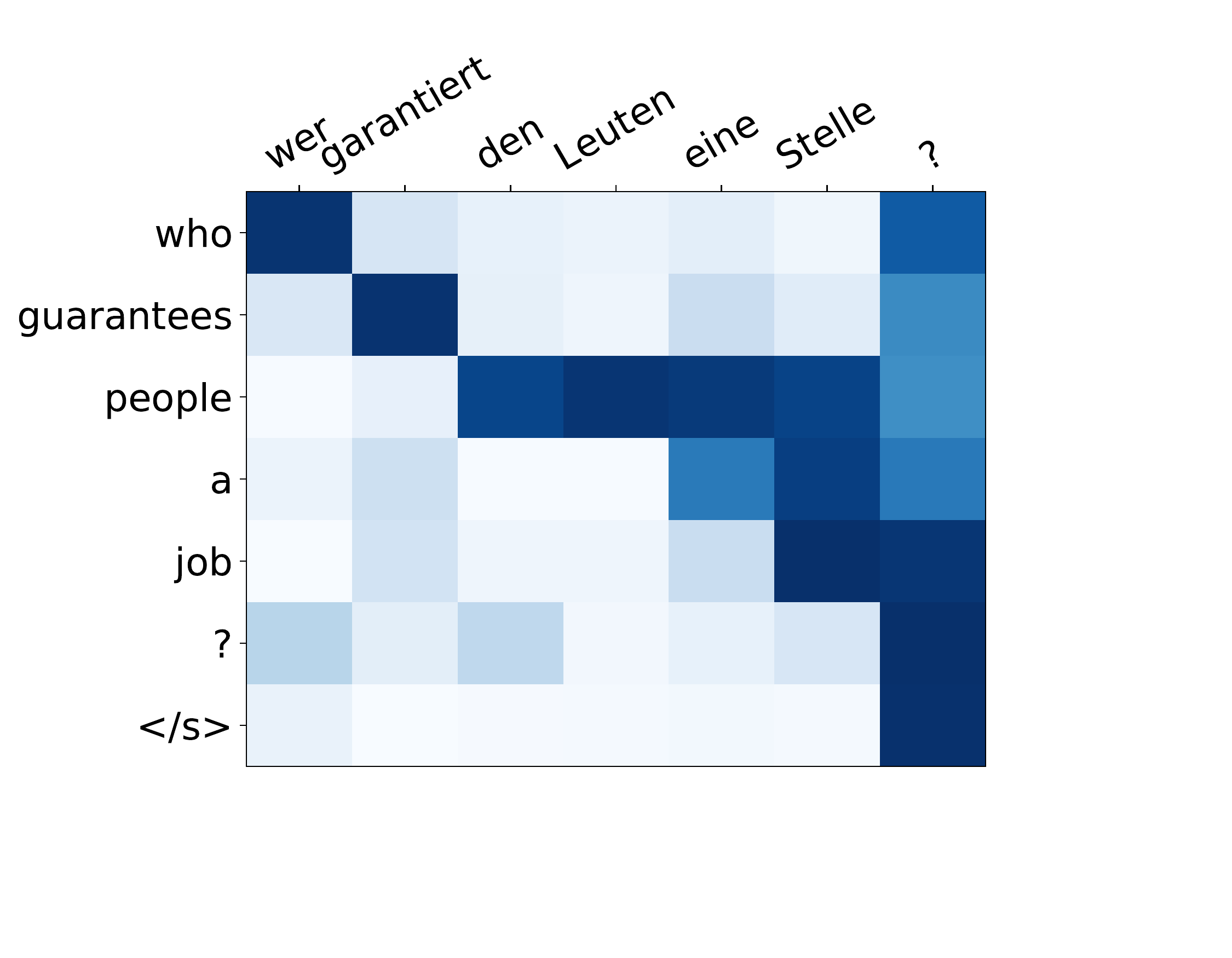} & 
\includegraphics[totalheight=3.17cm]{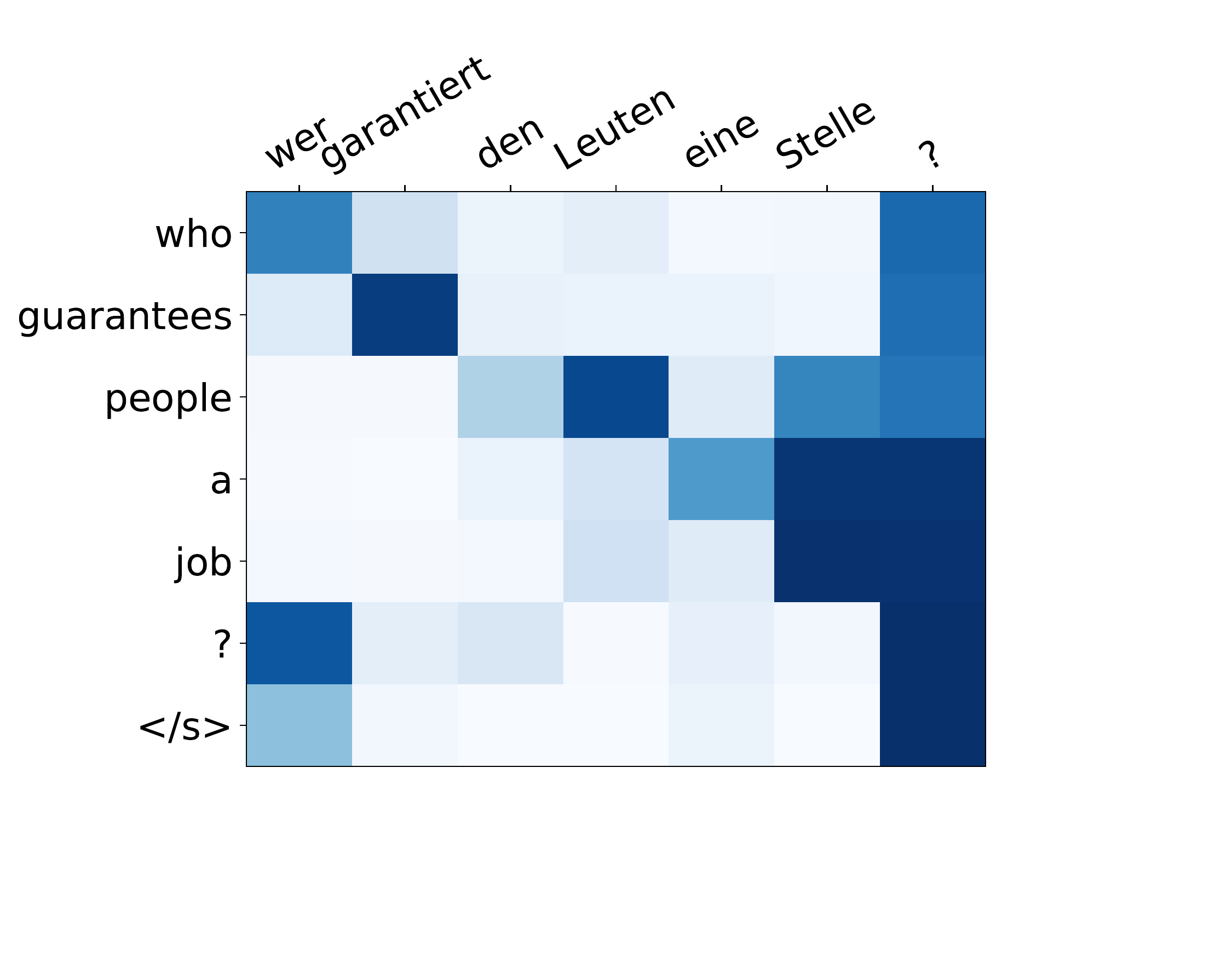} \\
\small (d) Layer 4 & \small (e) Layer 5 & \small (f) Layer 6\\ 
\includegraphics[totalheight=3.17cm]{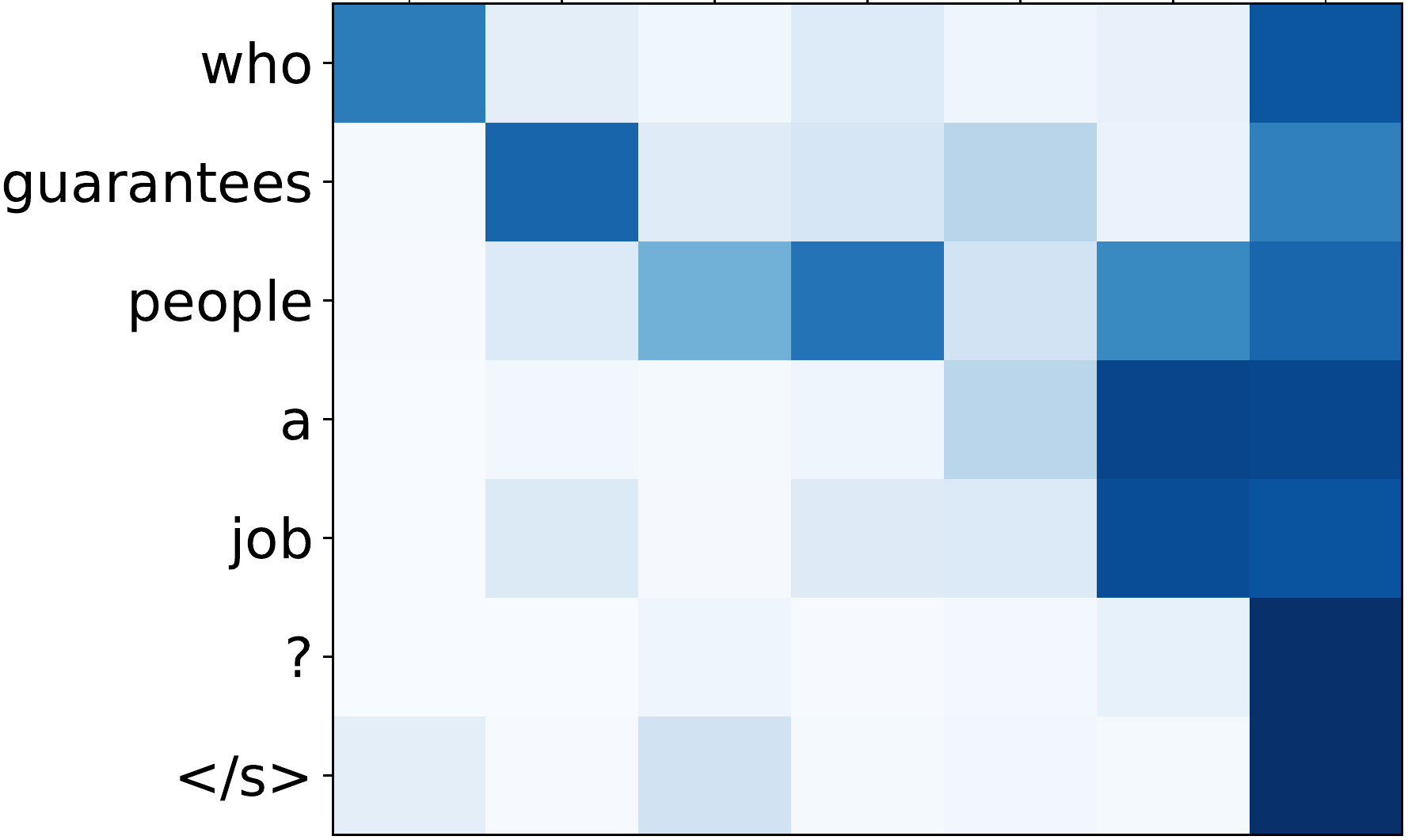} & 
\includegraphics[totalheight=3.17cm]{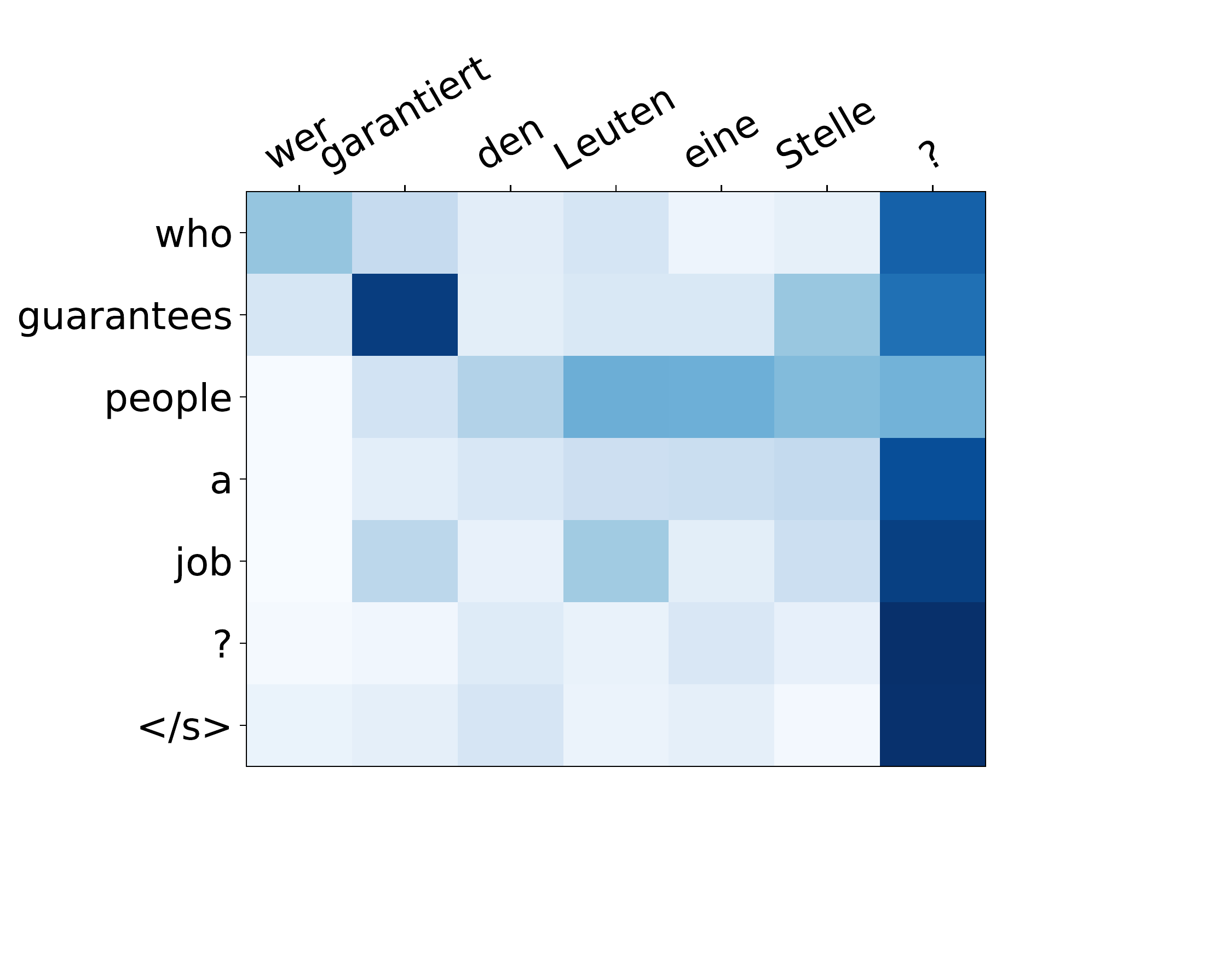} & 
\includegraphics[totalheight=3.17cm]{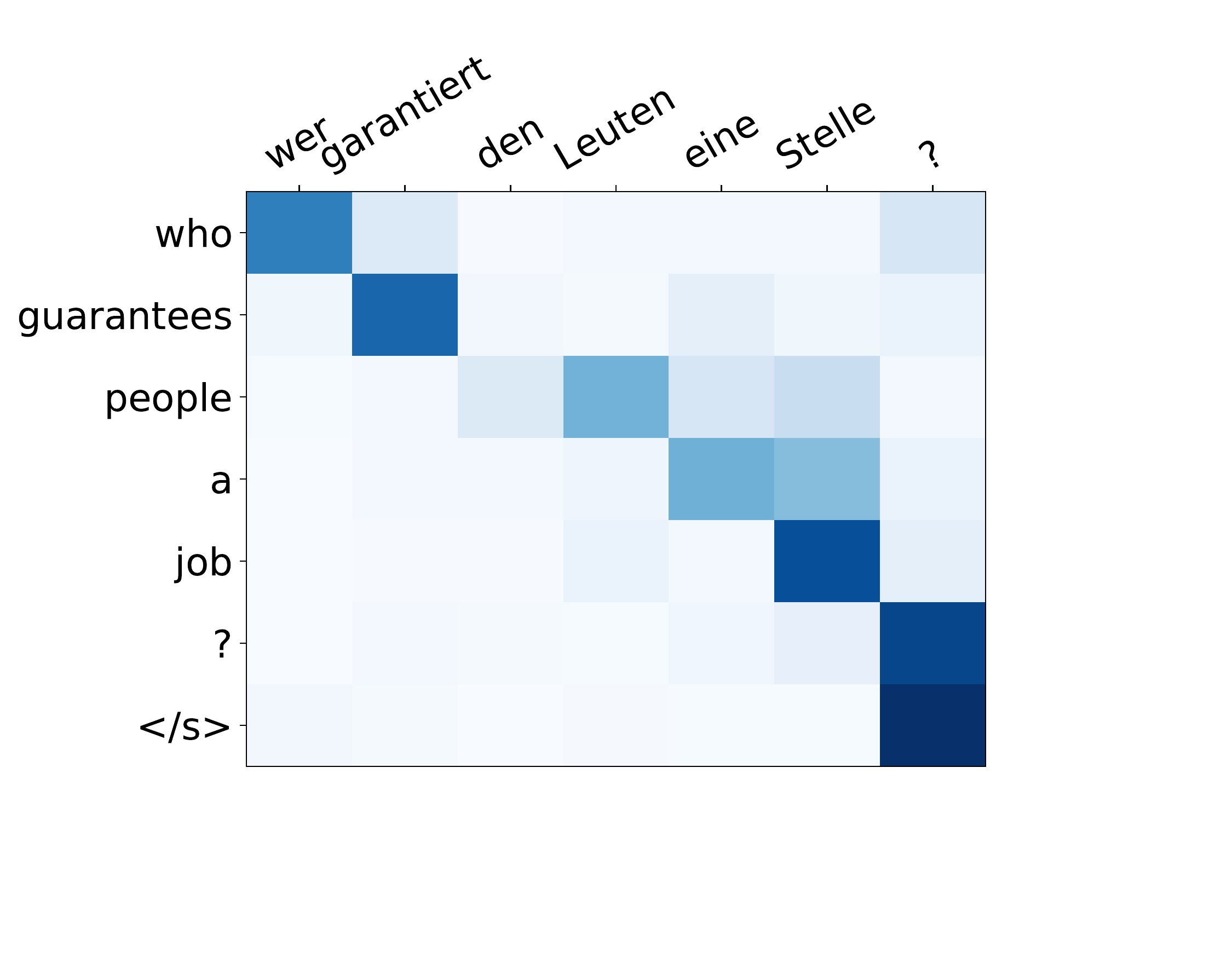} \\
\small (g) Layer 7 & \small (h) Layer 8 & \small (i) RNN \\
\end{tabular}
\caption{\label{table-attention} An example of attention visualization (German$\rightarrow$English). 
Each row is the attention distribution over all the source tokens at each time step. 
Each column represents the attention weight over a source token at all the time steps. 
Layer 1 to Layer 8 are attention layers in the Transformer model. 
Each attention layer has 8 heads, and the attention weights in each row are the 
maximal of all the heads. Thus, the summation of attention weights in each row is larger than 1. 
Darker blue means larger attention weights. }
\end{center}
\end{table*}
}

\noindent
We further validate the hypothesis by visualizing the attention distributions. 
Table \ref{table-attention} demonstrates the visualization of attention 
distributions of different attention mechanisms. 

`Stelle' is an ambiguous noun, whose reference translations are `job/position/work'. 
`Stelle' also has other translations such as `location/spot/site'. 
The context tokens `garantiert' (guarantee) and `Leuten' (people) contribute to disambiguating 
`Stelle'. However, the RNN model could translate `Stelle' correctly 
but only pays a little attention to `Leuten'. 

In the first layer, the attention mechanism does not pay attention to the correct 
source tokens if we only consider the larger attention weights. 
Then the ``alignment'' is learnt gradually in the following layers. 
The attention mechanism could pay attention to all the correct source tokens 
in the fifth layer. 
In addition, the attention mechanism could learn to pay attention to the related but 
unaligned source tokens in the eighth layer. 
For instance, the attention mechanism also attends to `Stelle' 
when generating `guarantees', and attends to `garantiert'  
and `Leuten' when generating `job'. 
These source tokens are not clearly attended to in the fifth layer. 

Since the vanilla attention mechanism is only one layer with one head, 
it does not perform as well as the advanced attention mechanism in learning to 
pay attention to context tokens. For instance, the attention mechanism in RNN 
only distributes a little attention to `Leuten' when generating `job'. 

\begin{figure}[htbp]
\centering
        \includegraphics[totalheight=5cm]{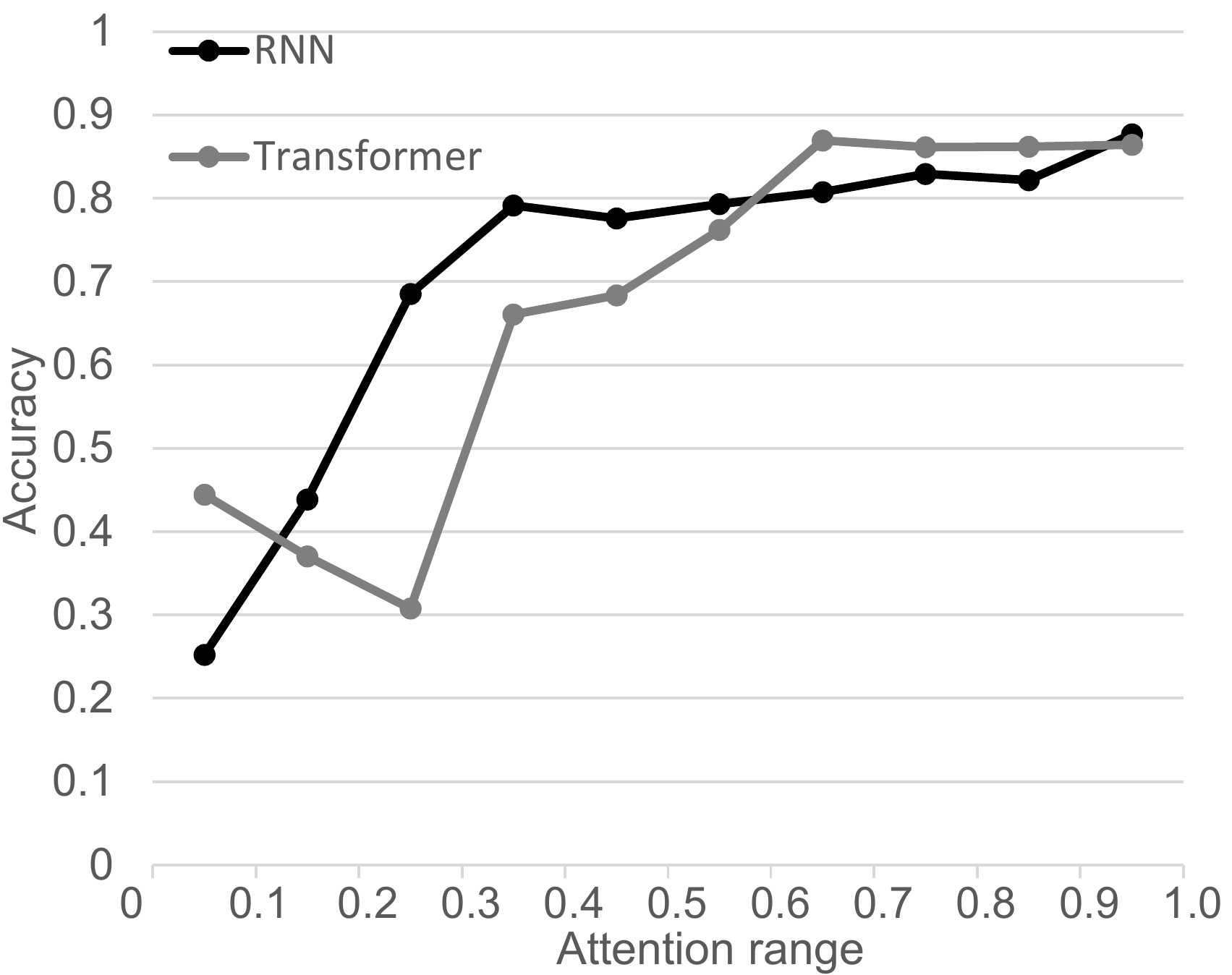}
    \caption{WSD accuracy over attention ranges.}
    \label{fig:acc-att}
\end{figure}

\subsection{Accuracy and Attention Weights}
\label{ssec:accuracy_and_weights}

We explore the relation between WSD accuracy and the attention weights over 
ambiguous nouns. As the alignment learnt by \textit{fast-align} does not guarantee 
that each ambiguous noun is aligned to the corresponding translation, we only 
consider the translations belonging to Group \textit{C1}, \textit{W1}, and \textit{Drop}. 
Figure \ref{fig:acc-att} shows the WSD accuracy over different attention ranges. 
Obviously, the accuracy is higher when the attention weight is greater. 
This result further confirms our assumption in Section \ref{sec:ambi} that 
the contextual information for disambiguation has been learnt by the encoder. 
In the attention range $(0, 0.3)$, the small attention weight causes many 
ambiguous nouns to be untranslated, which results in low WSD accuracy.

\subsection{Error Distribution}
\label{sec:error}

\begin{figure}[htbp]
\centering
        \includegraphics[totalheight=5.0cm]{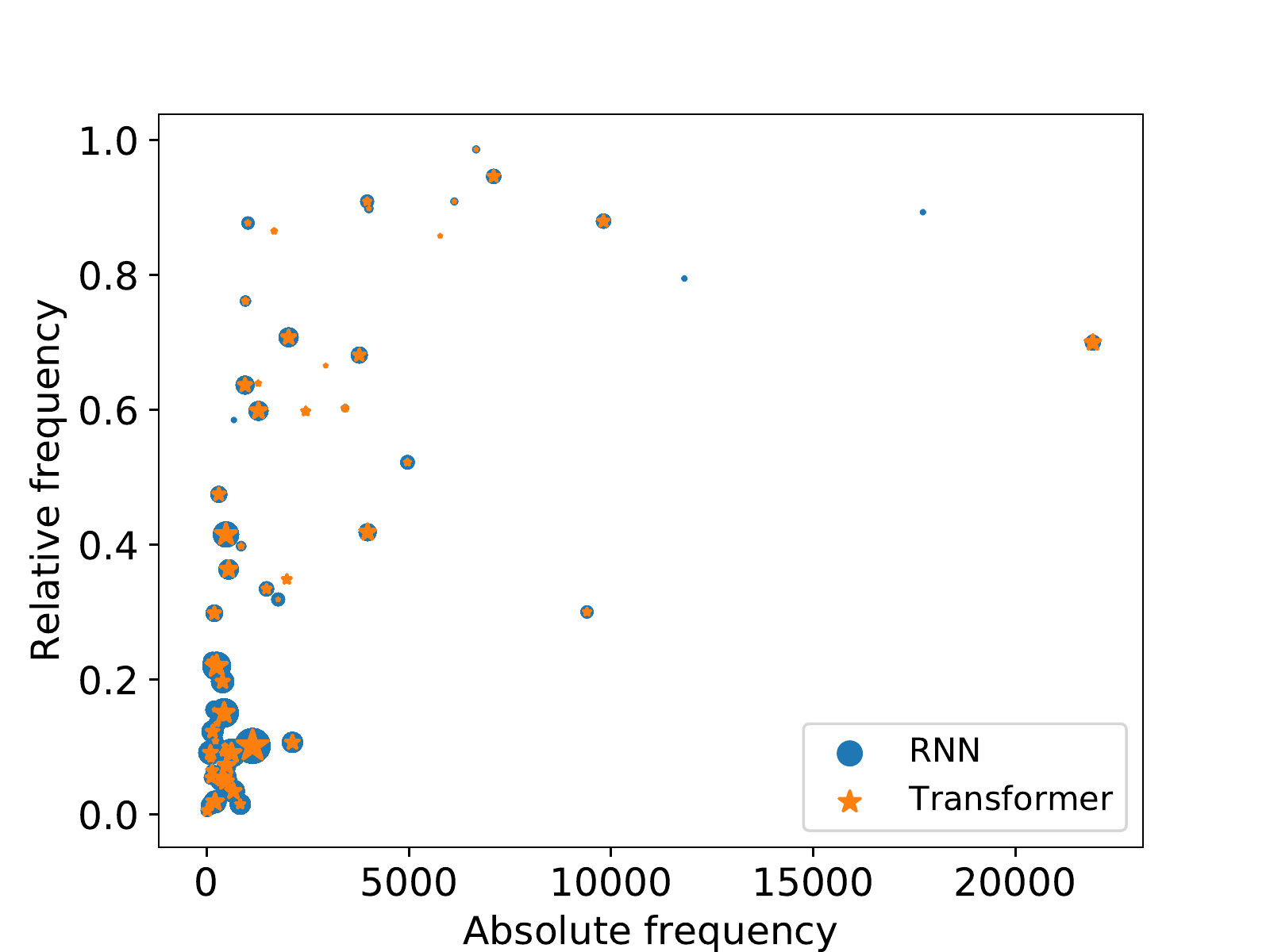}
    \caption{Error distributions over frequency. 
    Absolute frequency is the sense frequency in training set. 
    Relative frequency is the sense frequency in relation to source word frequency. 
    The size of the marker indicates how often the error occurs.}
    \label{fig:distri_error}
\end{figure}

Figure~\ref{fig:distri_error} shows the error distributions over absolute frequency 
(sense frequency in the training set) and relative frequency (sense frequency 
to source word frequency). The frequency information is given in the test set. 
It is very clear that most of the errors are in the left bottom corner 
which are low in both absolute frequency and relative frequency. 
There are 84.1\% and 80.8\% errors with an absolute frequency of less than 2000 
in \textit{RNN} and \textit{Transformer}, respectively. 

Even though the attention mechanism pays a lot of attention to a low-frequency sense, 
the model is still likely to generate an incorrect translation. 
Our evaluation method is different from \newcite{rios2017improving}, but the 
finding is the same, namely that data sparsity leads to incorrect translations.

\section{Conclusion} 
\label{sub:conclusion}

In this paper, we analyze two different attention mechanisms with respect to 
WSD in NMT. We evaluate the translations of ambiguous nouns directly rather than 
scoring the contrastive translations pairs, using \textit{ContraWSD} as the test set. 
We show that the WSD accuracy of these two models is around 80.2\% and 85.5\%, 
respectively. 
Data sparsity is the main problem causing incorrect translations. 
We hypothesize that attention mechanisms distribute more attention to context 
tokens to guide the translation of ambiguous nouns. However, we find that 
attention mechanisms are likely to pay more attention to the ambiguous noun itself.  
Compared to vanilla attention mechanisms, we reveal that the first few layers in 
Transformer attention mechanisms learn to ``align'' source and target tokens, 
while the last few layers learn to distribute attention to the related but unaligned 
context tokens. 
We conclude that attention mechanism is not the main mechanism used by NMT models to incorporate contextual information for WSD. 
In addition, Section \ref{ssec:accuracy_and_weights} has told us that the larger attention weights, the higher WSD accuracy. 
\newcite{Tang2018why} have shown that Transformer models are better than RNN models 
in WSD because of their stronger encoding ability. 
These results suggest that NMT models learn to encode contextual information necessary for WSD in the encoder hidden states. 


The question how NMT models learn to represent word senses and similar phenomena has implications for transfer learning, the diagnosis of translation errors, and for the design of architectures for MT, including architectures that scale up the context window to the level of documents. We hope that future work will continue to deepen our understanding of the internal workings of NMT models.

\section*{Acknowledgments}
We thank all the anonymous reviews who give a lot of valuable and insightful comments. 
Gongbo Tang is funded by the Chinese Scholarship Council (grant number \texttt{201607110016}). 

\bibliography{wsdnmt}

\begin{thebibliography}{24}
\expandafter\ifx\csname natexlab\endcsname\relax\def\natexlab#1{#1}\fi

\bibitem[{Bahdanau et~al.(2015)Bahdanau, Cho, and Bengio}]{bahdanau15joint}
Dzmitry Bahdanau, Kyunghyun Cho, and Yoshua Bengio. 2015.
\newblock \href {https://arxiv.org/pdf/1409.0473.pdf} {Neural machine
  translation by jointly learning to align and translate}.
\newblock In \emph{Proceedings of the 3rd International Conference on Learning
  Representations}, San Diego, USA.

\bibitem[{Chen et~al.(2015)Chen, Li, Li, Lin, Wang, Wang, Xiao, Xu, Zhang, and
  Zhang}]{chen2015mxnet}
Tianqi Chen, Mu~Li, Yutian Li, Min Lin, Naiyan Wang, Minjie Wang, Tianjun Xiao,
  Bing Xu, Chiyuan Zhang, and Zheng Zhang. 2015.
\newblock \href {http://arxiv.org/abs/1512.01274} {Mxnet: A flexible and
  efficient machine learning library for heterogeneous distributed systems}.
\newblock In \emph{Proceedings of the Workshop on Machine Learning Systems in
  Neural Information Processing Systems 2015}.

\bibitem[{Cho et~al.(2014)Cho, van Merrienboer, Gulcehre, Bahdanau, Bougares,
  Schwenk, and Bengio}]{cho2014learning}
Kyunghyun Cho, Bart van Merrienboer, Caglar Gulcehre, Dzmitry Bahdanau, Fethi
  Bougares, Holger Schwenk, and Yoshua Bengio. 2014.
\newblock \href {http://www.aclweb.org/anthology/D14-1179} {Learning phrase
  representations using {RNN} encoder--decoder for statistical machine
  translation}.
\newblock In \emph{Proceedings of the 2014 Conference on Empirical Methods in
  Natural Language Processing}, pages 1724--1734, Doha, Qatar. Association for
  Computational Linguistics.

\bibitem[{Domhan(2018)}]{Domhan2018how}
Tobias Domhan. 2018.
\newblock \href {http://aclweb.org/anthology/P18-1167} {How much attention do
  you need? a granular analysis of neural machine translation architectures}.
\newblock In \emph{Proceedings of the 56th Annual Meeting of the Association
  for Computational Linguistics (Volume 1: Long Papers)}, pages 1799--1808.
  Association for Computational Linguistics.

\bibitem[{Dyer et~al.(2013)Dyer, Chahuneau, and Smith}]{dyer2013fast}
Chris Dyer, Victor Chahuneau, and Noah~A. Smith. 2013.
\newblock \href {http://www.aclweb.org/anthology/N13-1073} {A simple, fast, and
  effective reparameterization of ibm model 2}.
\newblock In \emph{Proceedings of the 2013 Conference of the North American
  Chapter of the Association for Computational Linguistics: Human Language
  Technologies}, pages 644--648, Atlanta, USA. Association for Computational
  Linguistics.

\bibitem[{Gehring et~al.(2017)Gehring, Auli, Grangier, Yarats, and
  Dauphin}]{gehring2017convolutional}
Jonas Gehring, Michael Auli, David Grangier, Denis Yarats, and Yann~N. Dauphin.
  2017.
\newblock \href {http://proceedings.mlr.press/v70/gehring17a.html}
  {Convolutional sequence to sequence learning}.
\newblock In \emph{Proceedings of the 34th International Conference on Machine
  Learning}, pages 1243--1252, Sydney, Australia. The Proceedings of Machine
  Learning Research.

\bibitem[{Ghader and Monz(2017)}]{ghader2017what}
Hamidreza Ghader and Christof Monz. 2017.
\newblock \href {http://www.aclweb.org/anthology/I17-1004} {What does attention
  in neural machine translation pay attention to?}
\newblock In \emph{Proceedings of the Eighth International Joint Conference on
  Natural Language Processing (Volume 1: Long Papers)}, pages 30--39, Taipei,
  Taiwan. Asian Federation of Natural Language Processing.

\bibitem[{Hieber et~al.(2017)Hieber, Domhan, Denkowski, Vilar, Sokolov,
  Clifton, and Post}]{Hieber2017sockeye}
Felix Hieber, Tobias Domhan, Michael Denkowski, David Vilar, Artem Sokolov, Ann
  Clifton, and Matt Post. 2017.
\newblock \href {http://arxiv.org/abs/1712.05690} {Sockeye: A toolkit for
  neural machine translation}.
\newblock \emph{arXiv preprint arXiv:1712.05690}.

\bibitem[{Kalchbrenner and Blunsom(2013)}]{kal2013recurrent}
Nal Kalchbrenner and Phil Blunsom. 2013.
\newblock \href {http://www.aclweb.org/anthology/D13-1176} {Recurrent
  continuous translation models}.
\newblock In \emph{Proceedings of the 2013 Conference on Empirical Methods in
  Natural Language Processing}, pages 1700--1709, Seattle, USA. Association for
  Computational Linguistics.

\bibitem[{Kingma and Ba(2015)}]{Kingma2014AdamAM}
Diederik~P. Kingma and Jimmy Ba. 2015.
\newblock \href {https://arxiv.org/abs/1412.6980} {Adam: A method for
  stochastic optimization}.
\newblock In \emph{Proceedings of the 3rd International Conference on Learning
  Representations}, San Diego, California, USA.

\bibitem[{Koehn and Knowles(2017)}]{koehn2017challenges}
Philipp Koehn and Rebecca Knowles. 2017.
\newblock \href {http://www.aclweb.org/anthology/W17-3204} {Six challenges for
  neural machine translation}.
\newblock In \emph{Proceedings of the First Workshop on Neural Machine
  Translation}, pages 28--39, Vancouver, Canada. Association for Computational
  Linguistics.

\bibitem[{Koehn et~al.(2003)Koehn, Och, and Marcu}]{koehn2003statistical}
Philipp Koehn, Franz~Josef Och, and Daniel Marcu. 2003.
\newblock \href {http://aclweb.org/anthology/N/N03/N03-1017.pdf} {Statistical
  phrase-based translation}.
\newblock In \emph{Proceedings of the 2003 Conference of the North American
  Chapter of the Association for Computational Linguistics on Human Language
  Technology-Volume 1}, pages 48--54, Edmonton, Canada. Association for
  Computational Linguistics.

\bibitem[{Liu et~al.(2018)Liu, Lu, and Neubig}]{liu2017handling}
Frederick Liu, Han Lu, and Graham Neubig. 2018.
\newblock \href {http://aclweb.org/anthology/N18-1121} {Handling homographs in
  neural machine translation}.
\newblock In \emph{Proceedings of the 2018 Conference of the North American
  Chapter of the Association for Computational Linguistics: Human Language
  Technologies, Volume 1 (Long Papers)}, pages 1336--1345. Association for
  Computational Linguistics.

\bibitem[{Luong et~al.(2015)Luong, Pham, and Manning}]{luong2015effective}
Thang Luong, Hieu Pham, and Christopher~D. Manning. 2015.
\newblock \href {http://aclweb.org/anthology/D15-1166} {Effective approaches to
  attention-based neural machine translation}.
\newblock In \emph{Proceedings of the 2015 Conference on Empirical Methods in
  Natural Language Processing}, pages 1412--1421, Lisbon, Portugal. Association
  for Computational Linguistics.

\bibitem[{Marvin and Koehn(2018)}]{marvin2018exploring}
Rebecca Marvin and Phillip Koehn. 2018.
\newblock \href
  {https://amtaweb.org/wp-content/uploads/2018/03/AMTA_2018_Proceedings_Research_Track.pdf#page=131}
  {Exploring word sense disambiguation abilities of neural machine translation
  systems}.
\newblock In \emph{Proceedings of AMTA 2018 (Volume 1: MT Research Track)},
  pages 125--131, Boston, USA. Association for Machine Translation in the
  Americas.

\bibitem[{Miller(1995)}]{miller1995wordnet}
George~A Miller. 1995.
\newblock \href {https://dl.acm.org/citation.cfm?id=219748} {Wordnet: a lexical
  database for english}.
\newblock \emph{Communications of the ACM}, 38(11):39--41.

\bibitem[{Rios et~al.(2017)Rios, Mascarell, and Sennrich}]{rios2017improving}
Annette Rios, Laura Mascarell, and Rico Sennrich. 2017.
\newblock \href {http://aclweb.org/anthology/W/W17/W17-4702.pdf} {Improving
  word sense disambiguation in neural machine translation with sense
  embeddings}.
\newblock In \emph{Proceedings of the Second Conference on Machine
  Translation}, pages 11--19, Copenhagen, Denmark. Association for
  Computational Linguistics.

\bibitem[{Schmid(1999)}]{schmid1999treetagger}
Helmut Schmid. 1999.
\newblock \href {https://link.springer.com/chapter/10.1007/978-94-017-2390-9_2}
  {Improvements in part-of-speech tagging with an application to german}.
\newblock In \emph{Natural language processing using very large corpora}, pages
  13--25. Springer.

\bibitem[{Sennrich et~al.(2016)Sennrich, Haddow, and Birch}]{sennrich16sub}
Rico Sennrich, Barry Haddow, and Alexandra Birch. 2016.
\newblock \href {http://www.aclweb.org/anthology/P16-1162} {Neural machine
  translation of rare words with subword units}.
\newblock In \emph{Proceedings of the 54th Annual Meeting of the Association
  for Computational Linguistics (Volume 1: Long Papers)}, pages 1715--1725,
  Berlin, Germany. Association for Computational Linguistics.

\bibitem[{Sutskever et~al.(2014)Sutskever, Vinyals, and
  Le}]{sutskever2014sequence}
Ilya Sutskever, Oriol Vinyals, and Quoc~V Le. 2014.
\newblock \href
  {https://papers.nips.cc/paper/5346-sequence-to-sequence-learning-with-neural-networks}
  {Sequence to sequence learning with neural networks}.
\newblock In \emph{Proceedings of the Neural Information Processing Systems
  2014}, pages 3104--3112, Montr\'eal, Canada.

\bibitem[{Tang et~al.(2018{\natexlab{a}})Tang, Cap, Pettersson, and
  Nivre}]{Tang2018evaluation}
Gongbo Tang, Fabienne Cap, Eva Pettersson, and Joakim Nivre.
  2018{\natexlab{a}}.
\newblock \href {http://aclweb.org/anthology/C18-1112} {An evaluation of neural
  machine translation models on historical spelling normalization}.
\newblock In \emph{Proceedings of the 27th International Conference on
  Computational Linguistics}, pages 1320--1331. Association for Computational
  Linguistics.

\bibitem[{Tang et~al.(2018{\natexlab{b}})Tang, M{\"u}ller, Rios, and
  Sennrich}]{Tang2018why}
Gongbo Tang, Mathias M{\"u}ller, Annette Rios, and Rico Sennrich.
  2018{\natexlab{b}}.
\newblock \href {http://arxiv.org/abs/1808.08946} {{Why Self-Attention? A
  Targeted Evaluation of Neural Machine Translation Architectures}}.
\newblock In \emph{{Proceedings of the 2018 Conference on Empirical Methods in
  Natural Language Processing}}, Brussels, Belgium. Association for
  Computational Linguistics.

\bibitem[{Vaswani et~al.(2017)Vaswani, Shazeer, Parmar, Uszkoreit, Jones,
  Gomez, Kaiser, and Polosukhin}]{vaswani2017Attention}
Ashish Vaswani, Noam Shazeer, Niki Parmar, Jakob Uszkoreit, Llion Jones,
  Aidan~N Gomez, \L~ukasz Kaiser, and Illia Polosukhin. 2017.
\newblock \href
  {http://papers.nips.cc/paper/7181-attention-is-all-you-need.pdf} {Attention
  is all you need}.
\newblock In \emph{Advances in Neural Information Processing Systems 30}, pages
  6000--6010. Curran Associates, Inc.

\bibitem[{Vickrey et~al.(2005)Vickrey, Biewald, Teyssier, and
  Koller}]{vickrey2005word}
David Vickrey, Luke Biewald, Marc Teyssier, and Daphne Koller. 2005.
\newblock \href {http://www.aclweb.org/anthology/H/H05/H05-1097.pdf}
  {Word-sense disambiguation for machine translation}.
\newblock In \emph{Proceedings of Human Language Technology Conference and
  Conference on Empirical Methods in Natural Language Processing}, pages
  771--778, Vancouver, Canada. Association for Computational Linguistics.

\end{thebibliography}
\bibliographystyle{acl_natbib}

\end{document}